\documentclass[runningheads]{llncs}
\usepackage{graphicx}
\usepackage{amsmath,amssymb, bm} % define this before the line numbering.
\usepackage{caption}
\DeclareMathOperator*{\argmax}{arg\,max}
\DeclareMathOperator*{\argmin}{arg\,min}
\usepackage{color}
\usepackage{subfig}
\usepackage[width=122mm,left=12mm,paperwidth=146mm,height=193mm,top=12mm,paperheight=217mm]{geometry}
\usepackage{verbatim}
\usepackage[ruled]{algorithm2e}

\def\bal#1\eal{\begin{align}#1\end{align}} % align environment

\newcommand{\pr}[1]{\left(#1\right)} % left-right parenthesis 
\def\m{\mathbf}

\def\R{\mathbb{R}}

\newcommand{\norm}[2]{\ensuremath{\left\|#1\right\|_{#2}}}
\newcommand {\bbmtx}{\begin{bmatrix}} % begin matrix environment
\newcommand {\ebmtx}{\end{bmatrix}} % end matrix environment
\newcommand{\mean}[1]{\bar{#1}}
\usepackage[hidelinks]{hyperref}
\hypersetup{
  colorlinks   = true, %Colours links instead of ugly boxes
  urlcolor     = blue, %Colour for external hyperlinks
  linkcolor    = blue, %Colour of internal links
  citecolor   = green %Colour of citations
}
\usepackage{tikz}
\usetikzlibrary{spy, calc}

\newif\ifblackandwhitecycle

\begin{document}
% \renewcommand\thelinenumber{\color[rgb]{0.2,0.5,0.8}\normalfont\sffamily\scriptsize\arabic{linenumber}\color[rgb]{0,0,0}}
% \renewcommand\makeLineNumber {\hss\thelinenumber\ \hspace{6mm} \rlap{\hskip\textwidth\ \hspace{6.5mm}\thelinenumber}}
% \linenumbers
\pagestyle{headings}
\mainmatter

\title{Deep Image Demosaicking using a Cascade of Convolutional Residual Denoising Networks} 
% Replace with your title

\titlerunning{Deep Image Demosaicking with Residual Networks}
% Replace with a meaningful short version of your title

\authorrunning{F. Kokkinos and S. Lefkimmiatis}
% Replace with shorter version of the author list. If there are more authors than fits a line, please use A. Author et al.

\author{Filippos Kokkinos and Stamatios Lefkimmiatis\\ {\tt\small \{filippos.kokkinos, s.lefkimmiatis\}@skoltech.ru}}

%Please write out author names in full in the paper, i.e. full given and family names. 
%If any authors have names that can be parsed into FirstName LastName in multiple ways, please include the correct parsing, in a comment to the volume editors:
%\index{Lastnames, Firstnames}
%(Do not uncomment it, because you may introduce extra index items if you do that, we will use scripts for introducing index entries...)

\institute{Skolkovo Institute of Science and Technology (Skoltech), Moscow, Russia}

\maketitle

\begin{abstract}
Demosaicking and denoising are among the most crucial steps of modern digital camera pipelines and their joint treatment is a highly ill-posed inverse problem where at-least two-thirds of the information are missing and the rest are corrupted by noise. This poses a great challenge in obtaining meaningful reconstructions and a special care for the efficient treatment of the problem is required. While there are several machine learning approaches that have been recently introduced to deal with joint image demosaicking-denoising, in this work we propose a novel deep learning architecture which is inspired by powerful classical image regularization methods and large-scale convex optimization techniques. Consequently, our derived network is more transparent and has a clear interpretation compared to alternative competitive deep learning approaches. Our extensive experiments demonstrate that our network outperforms any previous approaches on both noisy and noise-free data. This improvement in reconstruction quality is attributed to the principled way we design our network architecture, which also requires fewer trainable parameters than the current state-of-the-art deep network solution. Finally, we show that our network has the ability to generalize well even when it is trained on small datasets, while keeping the overall number of trainable parameters low.

\end{abstract}

\section{Introduction}
Modern digital cameras perform a certain number of processing steps in order to create high quality images from raw sensor data. The sequence of the required processing steps is known as the imaging pipeline and the first two and most crucial steps involve image denoising and demosaicking. Both of these problems belong to the category of ill-posed problems while their joint treatment is very challenging since two-thirds of the underlying data are missing and the rest are perturbed by noise. It is clear that reconstruction errors during this early stage of the camera pipeline will eventually lead to unsatisfying final results. Furthermore, due to the modular nature of the camera processing pipelines, demosaicking and denoising were traditionally dealt in the past in a sequential manner. In detail, demosaicking algorithms reconstruct the image from unreliable spatially-shifted sensor data which introduce non-linear pixel noise, casting denoising an even harder problem. Since, demosaicking is an essential step of the camera pipeline, it has been extensively studied. For a complete survey of recent approaches, we refer to~\cite{766768}. One of the main drawbacks of several of the currently introduced methods that deal with the demosaicking problem, is that they assume a specific Bayer  pattern\cite{766768,zhang2011color,duran2014self,buades2009self,heide2014flexisp,chang2015color}. This is a rather strong assumption and limits their applicability since there are many cameras available in the market that employ different Color filter Array (CFA) patterns. Therefore, demosaicking methods that are able to generalize to different CFA patterns are preferred.

One simple method that works for any CFA pattern is bilinear interpolation on the neighboring values for a given pixel for each channel. The problem with this approach is the produced zippering artifacts which occur along high frequency signal changes, e.g., edges. Therefore, many approaches involve edge-adaptive interpolation schemes which follow the direction of the gradient of strong edges \cite{766768}.  However, the real challenges of demosaicking extend in the exploitation of both intra and inter-channel dependencies. The most common assumption is that color differences between color channels are constant, so that the end result leads to smooth images.  Other approaches make use of the self-similarity and redundancy properties of natural images \cite{buades2009self,zhang2011color,duran2014self,chang2015color}. Moreover, in some cases a post-processing step is applied to remove certain type of artifacts~\cite{Hirakawa.2005}. Another successful class consists of methods that act upon the frequency domain. Any Bayer CFA can be represented as the combination of a luminance component at baseband and two modulated components~\cite{alleysson.2005}. Upon this interpretation, Dubois \cite{dubois.2005,dubois.2006,dubois.2009} created a successful set of filter-banks using a least-squares method that was able to generalize to arbitrary sensor patterns. 

From the perspective of learning based approaches, the bibliography is short. A common problem with the design of learning based demosaicking algorithms is the lack of ground-truth images. In many approaches such as those  in~\cite{sun.2013,he2012self} the authors  used already processed images as references that are simulated mosaicked again, i.e. they apply a mosaick mask on the already demosaicked images, therefore obtaining non-realistic pairs for tuning trainable methods. In a recent work Khasabi et. al. \cite{khasabi2014} provided a way to produce a dataset with realistic reference images allowing for the design of machine learning demosaicking algorithms. We use the produced Microsoft Demosaicking dataset (MSR) \cite{khasabi2014} in order to train, evaluate and compare our system. The contained images have to be demosaicked in the linear RGB (linRGB) color space before being transformed via color transformation and gamma correction into standard RGB (sRGB) space. Furthermore, two common CFA patterns are contained into the dataset, namely Bayer and Fuji X Trans which enables the development and evaluation of methods that are able to deal with different CFA patterns.

Apart from the demosaicking problem, another problem that requires special attention is the elimination of noise arising from the sensor and which distorts the acquired raw data. Firstly, the sensor readings are corrupted with \textit{shot} noise~\cite{foi.2008} which is the result of random variation of the detected photons. Second, electronic inefficiencies during reading and converting electrical charge into a digital count exhibit another type of noise, namely \textit{read} noise. Under certain circumstances both noises can be approximated by noise following a heteroscedastic Gaussian pdf~\cite{foi.2008}. Prior work from Kalevo and Rantanen \cite{kalevo.2002}, analyzed whether denoising should occur before or after the demosaicking step. It was experimentally confirmed that denoising is preferably done before demosaicking. However, the case of joint denoising and demosaicking was not analyzed. In later work, many researchers \cite{menon.2009,zhang.2009,klatzer2016} showed that joint denoising and demosaicking yields better results. Motivated by these works, we also pursue a joint approach for denoising and demosaicking of raw sensor data. 

In a very recent work Gharbi et. al.~\cite{Gharbi:2016:DJD:2980179.2982399} exploit the advantages in the field of deep learning to create a Convolutional Neural Network (CNN) that is able to jointly denoise and demosaick images. Apart from the design of the aforementioned network, a lot of effort was put by the authors to  create a new large demosaicking dataset, namely the MIT Demosaicking Dataset which consists of 2.6 million patches of images. These patches were mined from a large collection of data following specific visual distortion metrics.

Our main contribution is a novel deep neural network for solving the joint image demosaicking-denoising problem\footnote{The code for both training and inference will be made available from the authors' website.}. The network architecture is inspired by classical image regularization approaches and a powerful optimization strategy that has been successfully used in the past for dealing with general inverse imaging problems. We demonstrate through extensive experimentation that our approach leads to higher-quality reconstruction than other competing methods in both linear RGB (linRGB) and standard RGB (sRGB) color spaces. Moreover, we further show that our derived network not only outperforms the current CNN-based state-of-the art network~\cite{Gharbi:2016:DJD:2980179.2982399}, but it achieves this by using less trainable parameters and by being trained only on a small fraction of the training data.

\section{Problem Formulation}
To solve the joint demosaicking-denoising problem, one of the most frequently used approaches in the literature relies on the following linear observation model
\begin{equation}
\label{eq:linearmodel}
\m y = \m M \m x+ \m n,
\end{equation}
which relates the observed sensor raw data, $\m y \in \R^N$, and the underlying image $\m x \in \R^N$ that we aim to restore. Both $\m x$ and $\m y$ correspond to the vectorized forms of the images assuming that they have been raster scanned using a lexicographical order. Under this notation, $\m M\in \R^{N \times N}$ is the degradation matrix that models the spatial response of the imaging device, and in particular the CFA pattern. According to this, $\m M$ corresponds to a square diagonal binary matrix where the zero elements in its diagonal indicate the spatial and channel locations in the image where color information is missing. Apart from the missing color values, the image measurements are also perturbed by noise which hereafter, we will assume that is an i.i.d Gaussian noise $\m n \sim \mathcal{N}(0,\,\sigma^2)$. Note, that this is a rather simplified assumption about the noise statistics distorting the measurements. However, this model only serves as our starting point based on which we will design our network architecture. In the sequel, our derived network will be trained and evaluated on images that are distorted by noise which follows statistics that better approximate real noisy conditions.

Recovering $\m x$ from the measurements $\m y$ belongs to the broad class of linear inverse problems. For the problem under study, the operator $\m M$ is clearly singular. This fact combined with the presence of noise perturbing the measurements leads to an ill-posed problem where a unique solution does not exist. One popular way to deal with this, is to adopt a Bayesian approach and seek for the Maximum A Posteriori (MAP) estimator
\begin{equation}
\label{eq:map}
{\m x}^\star = \argmax_{\m x} \log(p(\m x|\m y)) = \argmax_{\m x} \log(p(\m y| \m x)) + \log(p(\m x)),
\end{equation}
where $\log(p(\m y| \m x))$ represents the log-likelihood of the observation $\m y$ and $\log(p( \m x))$ represents the log-prior of $\m{x}$. Problem~\eqref{eq:map} can be equivalently re-casted as the minimization problem 
\begin{equation}
\label{eq:var}
{\m x}^\star = \argmin_{\m x} \frac{1}{2\sigma^2} \norm{\m y-\m M \m x}{2}^2 + \phi(\m x)
\end{equation}
where the first term corresponds to the negative log-likelihood (assuming i.i.d Gaussian noise of variance $\sigma ^2$) and the second term corresponds to the negative log-prior. According to the above, the restoration of the underlying image $\m x$, boils down to computing the minimizer of the objective function in Eq.~\eqref{eq:var}, which  consists of two terms. This problem formulation has also direct links to variational methods where the first term can be interpreted as the data-fidelity that quantifies the proximity of the solution to the observation and the second term can be seen as the regularizer, whose role is to promotes solutions that satisfy certain favorable image properties. 

In general, the minimization of the objective function 
\begin{equation}
\label{eq:variational}
Q(\m x)= \frac{1}{2\sigma^2}\norm{\m y-\m M \m x}{2}^2 + \phi(\m x)
\end{equation}
is far from a trivial task, especially when the function $\phi(\m x)$ is not of a quadratic form, which implies that the solution cannot simply be obtained by solving a set of linear equations. From the above, it is clear that there are two important challenges that need to be dealt with before we are in position of deriving a satisfactory solution for our problem. The first one is to come up with an algorithm that can efficiently minimize $Q\pr{\m x}$, while the second one is to select an appropriate form for $\phi\pr{\m x}$, which will constrain the set of admissible solutions by promoting only those that exhibit the desired properties. 

In Section~\ref{sec:MM}, we will focus on the first challenge, while in Section~\ref{sec:ResDNet} we will discuss how it is possible to avoid making any explicit decisions for the regularizer (or equivalently the negative log-prior) by following a machine learning approach. Such an approach will allow us to infer the form of $\phi\pr{\m x}$, in an indirect way, from training data. 

\section{Majorization-Minimization Framework}\label{sec:MM}
One of the main difficulties in the minimization of the objective function in Eq.~\eqref{eq:variational} is the coupling that exists between the singular degradation operator, $\m M$, and the latent image $\m x$. To circumvent this difficulty there are several optimization strategies available that we could rely on, with potential candidates being splitting variables techniques such as the Alternating Direction Method of Multipliers~\cite{boyd.2011} and the Split Bregman approach~\cite{goldstein2009split}. However, one difficulty that arises by using such methods is that they involve additional parameters that need to be tuned so that a satisfactory convergence speed to the solution is achieved. Unfortunately, there is not a simple and straightforward way to choose these parameters. For this reason, in this work we will instead pursue a majorization-minimization (MM) approach~\cite{hunter2004tutorial,figueiredo2007majorization}, which does not pose such a requirement. Under this framework, as we will describe in detail, instead of obtaining the solution by minimizing \eqref{eq:variational}, we compute it iteratively via the successive minimization of surrogate functions. The surrogate functions provide an upper bound of the initial objective function \cite{hunter2004tutorial} and they are simpler to deal with than the original objective function. 

Specifically, in the majorization-minimization (MM) framework, an iterative algorithm for solving the minimization problem 
\begin{equation}
{\m x}^* = \argmin_f Q\pr{\m x}
\end{equation}
takes the form
\begin{equation} \label{eq:mm_iter}
\m x^{(t+1)} = \argmin_x \tilde{Q}(\m x;{\m x}^{(t)}),
\end{equation}
where $\tilde{Q}(\m x;{\m x}^{(t)})$ is the majorizer of the function $Q(\m x)$ at a fixed point ${\m x}^{(t)}$, satisfying the two conditions
\begin{equation}
\label{eq:prop_one}
\tilde{Q}(\m x;\m x^{(t)}) > Q(\m x), \forall \m x\ne \m x^{(t)}\quad \mbox{and}\quad 
\tilde{Q}(\m x^{(t)};\m x^{(t)}) = Q(\m x^{(t)}).
\end{equation}

Here, the underlying idea is that instead of minimizing the actual objective function ${Q}(\m x)$, we fist upper-bound it by a suitable majorizer $\tilde{Q}(\m x;{\m x}^{(t)})$, and then minimize this majorizing function to produce the next iterate $\m x^{(t+1)}$. Given the properties of the majorizer, iteratively minimizing $\tilde{Q}(\cdot;{\m x}^{(t)})$ also decreases the objective function $Q(\cdot)$. In fact, it is not even required that the surrogate function in each iteration is minimized, but it is sufficient to only find a $\m x^{(t+1)}$ that decreases it. 

To derive a majorizer for $Q\pr{\m x}$ we opt for a majorizer of the data-fidelity term (negative log-likelihood). In particular, we consider the following majorizer
\begin{equation}
\label{eq:majorizer}
\tilde{d}(\m x,{\m x}_0) = \frac{1}{2\sigma^2}\norm{\m y-\m M \m x}{2}^2 + d(\m x,\m x_0),
\end{equation}
where $d(\m x, \m x_0) = \frac{1}{2\sigma^2}(\m x-\m x_0)^T [ \alpha \m I- \m M^T \m M ](\m x- \m x_0)$ is a function that measures the distance between $\m x$ and $\m x_0$. Since $\m M$ is a binary diagonal matrix, it is an idempotent matrix, that is $\m M^{T} \m M= \m M$, and thus $d(\m x,\m x_0) = \frac{1}{2\sigma^2}(\m x- \m x_0)^T [ \alpha \m I- \m M ](\m x- \m x_0)$. According to the conditions in~\eqref{eq:prop_one}, in order $\tilde{d}(\m x, \m x_0)$ to be a valid majorizer,  we need to ensure that $d(\m x, \m x_0) \ge 0, \forall \m x$ with equality iff $\m x= \m x_0$. This suggests that $a \m I- \m M$ must be a positive definite matrix, which only holds when $\alpha > \norm{\m M}{2} = 1$, i.e $\alpha$ is bigger than the maximum eigenvalue of $ \m M$. Based on the above, the upper-bounded version of \eqref{eq:variational} is finally written as

\begin{equation}
\label{eq:upperboundeq}
\tilde{Q}(\m x,\m x_0) = \frac{1}{2(\sigma/\sqrt{a})^2} \norm{\m x- \m z}{2}^2 + \phi(\m x) + c,
\end{equation}
where c is a constant and $\m z = \m y + (\m I - \m M)\m x_0$. 

Notice that following this approach, we have managed to completely decouple the degradation operator $\m M$ from $\m x$ and we now need to deal with a simpler problem. In fact, the resulting surrogate function in Eq.~\eqref{eq:upperboundeq} can be interpreted as the objective function of a denoising problem, with $\m z$ being the noisy measurements that are corrupted by noise whose variance is equal to $\sigma^2/a$. This is a key observation that we will heavily rely on in order to design our deep network architecture. In particular, it is now possible instead of selecting the form of $\phi\pr{\m x}$ and minimizing the surrogate function, to employ a denoising neural network that will compute the solution of the current iteration. Our idea is similar in nature to other recent image restoration approaches that have employed denoising networks as part of alternative iterative optimization strategies, such as RED~\cite{romano2017little} and $P^3$~\cite{Venkatakrishnan.2013}. 
%However, the main difference of our approach with the aforementioned strategies is that our scheme does not involve any `free' parameter that needs to be carefully tuned.
This direction for solving the joint denoising-demosaicking problem is very appealing since by using training data we can implicitly learn the function $\phi\pr{\m x}$ and also minimize the corresponding surrogate function using a feed-forward network. This way we can completely avoid making any explicit decision for the regularizer or relying on an iterative optimization strategy to minimize the function in Eq.~\eqref{eq:upperboundeq}.

\section{Residual Denoising Network (ResDNet)}\label{sec:ResDNet}
\begin{figure}[t]%
	\centering
	\includegraphics[height=1.2in]{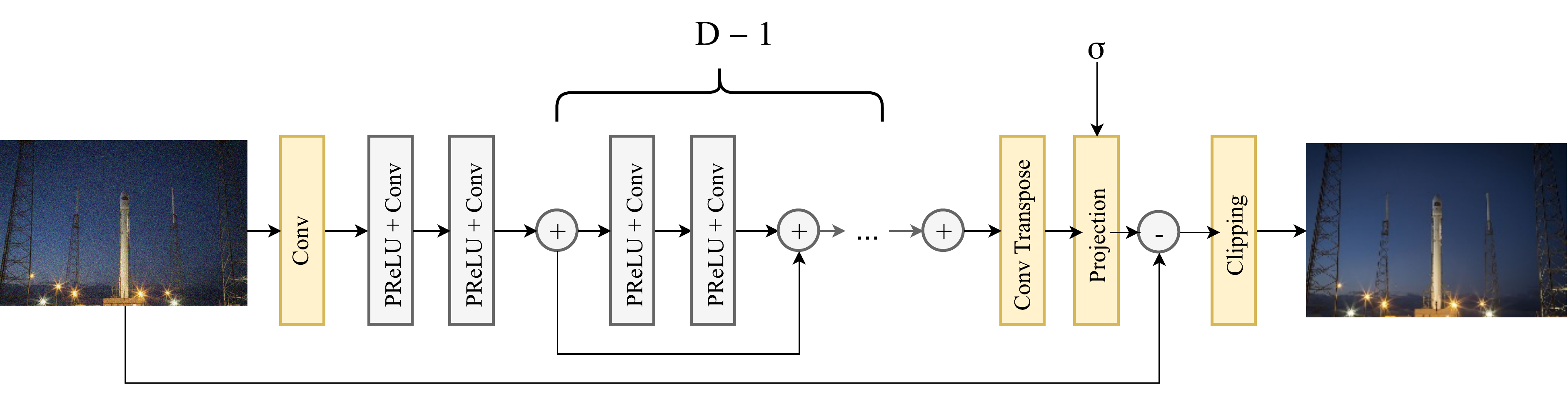}%
    \caption{The architecture of the proposed ResDNet denoising network, which serves as the back-bone of our overall system.}
\label{fig:network}
\end{figure}

Based on the discussion above, the most important part of our approach is the design of a denoising network that will play the role of the solver for the surrogate function in Eq.~\eqref{eq:upperboundeq}. The architecture of the proposed network is depicted in Fig.~\ref{fig:network}. This is a residual network similar to DnCNN~\cite{DCNN}, where the output of the network is subtracted from its input. Therefore, the network itself acts as a noise estimator and its task is to estimate the noise realization that distorts the input. Such network architectures have been shown to lead to better restoration results than alternative approaches~\cite{DCNN,lefkimmiatis.2017}. One distinctive difference between our network and DnCNN, which also makes our network suitable to be used as a part of the MM-approach, is that it accepts two inputs, namely the distorted input and the variance of the noise. This way, as we will demonstrate in the sequel, we are able to learn a single set of parameters for our network and to employ the same network to inputs that are distorted by a wide range of noise levels. While the blind version of DnCNN can also work for different noise levels, as opposed to our network it features an internal mechanism to estimate the noise variance. However, when the noise statistics deviate significantly from the training conditions such a mechanism can fail and thus DnCNN can lead to poor denoising results~\cite{lefkimmiatis.2017}. In fact, due to this reason in~\cite{zhang.2017}, where more general restoration problems than denoising are studied, the authors of DnCNN use a non-blind variant of their network as a part of their proposed restoration approach. Nevertheless, the drawback of this approach is that it requires the training of a deep network for each noise level. This can be rather impractical, especially in cases where one would like to employ such networks on devices with limited storage capacities. In our case, inspired by the recent work in~\cite{lefkimmiatis.2017} we circumvent this limitation by explicitly providing as input to our network the noise variance, which is then used to assist the network so as to provide an accurate estimate of the noise distorting the input. Note that there are several techniques available in the literature that can provide an estimate of the noise variance, such as those described in~\cite{Foi2009,Liu2013}, and thus this requirement does not pose any significant challenges in our approach.

A ResDNet with depth $D$, consists of five fundamental blocks. The first block is a convolutional layer with 64 filters whose kernel size is $5\times 5$. The second one is a non-linear block that consists of a parametrized rectified linear unit activation function (PReLU), followed by a convolution with 64 filters of $3\times 3$ kernels. The PReLU function is defined as $\text{PReLU}(\m x) = \max(0, \m x) + \bm \kappa * \min(0,\m x)$ where $\bm \kappa$ is a vector whose size is equal to the number of input channels. In our network we use $D*2$ distinct non-linear blocks  which we connect via a shortcut connection every second block in a similar manner to~\cite{he.2015} as shown in Fig.~\ref{fig:network}. Next, the output of the non-linear stage is processed by a transposed convolution layer which reduces the number of channels from 64 to 3 and has a kernel size of $5\times 5$. Then, it follows a projection layer~\cite{lefkimmiatis.2017} which accepts as an additional input the noise variance and whose role is to normalize the noise realization estimate so that it will have the correct variance, before this is subtracted from the input of the network. Finally the result is clipped so that the intensities of the output lie in the range $[0, 255]$. This last layer enforces our prior knowledge about the expected range of valid pixel intensities.

Regarding implementation details, before each convolution layer the input is padded to make sure that each feature map has the same spatial size as the input image. However, unlike the common approach followed in most of the deep learning systems for computer vision applications, we use reflexive padding than zero padding. Another important difference to other networks used for image restoration tasks~\cite{DCNN,zhang.2017} is that we don't use batch normalization after convolutions. Instead, we use the parametric convolution representation that has been proposed in~\cite{lefkimmiatis.2017} and which is motivated by image regularization related arguments. In particular, if $\m v\in \R^L$ represents the weights of a filter in a convolutional layer, these are parametrized as
\bal
\m v = \frac{s\pr{\m u -\mean{\m u}}}{\norm{\m u -\mean{\m u}}{2}},
\eal
where $s$ is a scalar trainable parameter, $\m u\in \R^L$ and $\mean{\m u}$ denotes the mean value of $\m u$. In other words, we are learning zero-mean valued filters whose $\ell_2$-norm is equal to $s$.

Furthermore, the projection layer, which is used just before the subtraction operation with the network input, corresponds to the following $\ell_2$ orthogonal projection 
\begin{equation}
\mathcal{P_{\mathcal{C}}}\pr{\m y} =\varepsilon \frac{\m y}{max(\norm{\m y}{2},\varepsilon)},
\end{equation}
where $\varepsilon = e^\gamma\theta$, $\theta=\sigma\sqrt{N-1}$, $N$ is the total number of pixels in the image (including the color channels), $\sigma$ is the standard deviation of the noise distorting the input, and $\gamma$ is a scalar trainable parameter. As we mentioned earlier, the goal of this layer is to normalize the noise realization estimate so that it has the desired variance before it is subtracted from the network input. 

\section{Demosaicking Network Architecture}
The overall architecture of our approach is based upon the MM framework, presented in Section~\ref{sec:MM}, and the proposed denoising network. As discussed, the MM is an iterative algorithm Eq.~\eqref{eq:mm_iter} where the minimization of the majorizer in Eq.~\eqref{eq:upperboundeq} can be interpreted as a denoising problem. One way to design the demosaicking network would be to unroll the MM algorithm as $K$ discrete steps and then for each step use a different denoising network to retrieve the solution of Eq.~\eqref{eq:upperboundeq}. However, this approach can have two distinct drawbacks which will hinder its performance. The first one, is that the usage of a different denoising neural network for each step like in \cite{zhang.2017}, demands a high overall number of parameters, which is equal to $K$ times the parameters of the employed denoiser, making the demosaicking network impractical for any real applications. To override this drawback, we opt to use our ResDNet denoiser, which can be applied to a wide range of noise levels, for all $K$ steps of our demosaick network, using the same set of parameters. By sharing the parameters of our denoiser across all the $K$ steps, the overall demosaicking approach maintains a low number of required parameters.

\begin{algorithm}[t]
 \SetAlgoCaptionSeparator{\unskip:}
 \SetKwInOut{Input}{Input}
 \Input{$\m M$ : CFA, $\m y$ : input, $K$ : iterations, $\bm w\in\R^K$ : extrapolation weights, $\bm \sigma\in\R^K$ : noise vector}
 $\m x^{0}= \m 0$, $\m x^{1}= \m y$\;
  \For{$i\gets1$ \KwTo $K$}{
    $ \m u = \m x ^{(i)} + \bm w_i (\m x^{(i)} - \m x^{(i-1)}) $\;
    $\m x^{(i+1)} = \text{ResDNet}((\m I-\m M) \m u + \m y , \bm \sigma_i )$\; 
    }
 \caption{The proposed demosaicking network described as an iterative process. The ResDnet parameters remain the same in every iteration.}
  \label{alg:the_alg}
\end{algorithm}

The second drawback of the MM framework as described in Section~\ref{sec:MM} is the slow convergence \cite{fista.2009} that it can exhibit. Beck and Teboulle \cite{fista.2009} introduced an accelerated version of this MM approach which combines the solutions of two consecutive steps with a certain extrapolation weight that is different for every step. In this work, we adopt a similar strategy which we describe in Algorithm~\ref{alg:the_alg}. Furthermore, in our approach we go one step further and instead of using the values originally suggested in~\cite{fista.2009} for the weights $\bm w \in \mathbb{R}^K$, we treat them as trainable parameters and learn them directly from the data. These weights are initialized with $w_i = \frac{i-1}{i+2}\text{,} \forall 1 \le i \le K$.

The convergence of our framework can be further sped up by employing a continuation strategy \cite{Lin2015} where the main idea is to solve the problem in Eq.~\eqref{eq:upperboundeq} with a large value of $\sigma$  and then gradually decrease it until the target value is reached.  Our approach is able to make use of the continuation strategy due to the design of our ResDNet denoiser, which accepts as an additional argument the noise variance. In detail, we initialize the trainable vector $\bm \sigma \in \mathbb{R}^K$ with values spaced evenly on a log scale from $\sigma_{max}$ to $\sigma_{min}$ and later on the vector $\bm \sigma$ is further finetuned on the training dataset by back-propagation training.

In summary, our overall demosaicking network is described in  Algorithm~\ref{alg:the_alg} where the set of trainable parameters $\theta$ consists of the parameters of the ResDNet denoiser, the extrapolation weights $\bm w$ and the noise level $\bm \sigma$. All of the aforementioned parameters are initialized as described in the current section and Section~\ref{sec:ResDNet} and are trained on specific demosaick datasets. In order to speed up the learning process, the employed ResDNet denoiser is pre-trained for a denoising task where multiple noise levels are considered.

Finally, while our demosaick network shares a similar philosophy with methods such as RED~\cite{romano2017little}, $P^3$~\cite{Venkatakrishnan.2013} and IRCNN~\cite{zhang.2017}, it exhibits some important and distinct differences. In particular, the aforementioned strategies make use of certain optimization schemes to decompose their original problem into subproblems that are solvable by a denoiser. For example, the authors of $P^3$~\cite{Venkatakrishnan.2013} decompose the original problem Eq.~\eqref{eq:linearmodel} via ADMM~\cite{boyd.2011} optimization algorithm and solve instead a linear system of equations and a denoising problem, where the authors of RED~\cite{romano2017little} go one step further and make use of the Lagrangian on par with a denoiser. Both approaches are similar to ours, however their formulation involves a tunable variable $\lambda$ that weights the participation of the regularizer on the overall optimization procedure. Thus, in order to obtain an accurate reconstruction in reasonable time, the user must manually tune the variable $\lambda$ which is not a trivial task. On the other hand, our method does not involve any tunable variables by the user. Furthermore, the approaches $P^3$, RED  and IRCNN are based upon static denoisers like Non Local Means~\cite{buades2005non}, BM3D~\cite{dabov2007image} and DCNN~\cite{DCNN}, meanwhile we opt to use a universal denoiser, like ResDnet, that can be further trained on any available training data. Finally, our approach goes one step further and we use a trainable version of an iterative optimization strategy for the task of the joint denoising-demosaicking in the form of a feed-forward neural network.

\section{Network Training}
\subsection{Image Denoising} \label{sec:resdnet_train}
The denoising network ResDnet that we use as part of our overall network is pre-trained on the Berkeley segmentation dataset (BSDS) \cite{937655}, which consists of 500 color images. These images were split in two sets, 400 were used to form a train set and the rest 100 formed a validation set. All the images were randomly cropped into patches of size $180 \times 180$ pixels. The patches were perturbed with noise $\sigma \in [0,15]$ and the network was optimized to minimize the Mean Square Error. We set the network depth $D=5$, all weights are initialized as in He et al. \cite{he.delving} and the optimization is carried out using ADAM \cite{kingma2014adam} which is a stochastic gradient descent algorithm which adapts the learning rate per parameter. The training procedure starts with an initial learning rate equal to $10^{-2}$.

\subsection{Joint Denoising and Demosaicking}
Using the pre-trained denoiser \ref{sec:resdnet_train}, our novel framework is further trained in an end-to-end fashion to minimize the averaged $L_1$ loss over a minibatch of size $d$,
\begin{equation}
L(\theta) = \frac{1}{N} \sum_{i=1}^d \norm{\m y_i - f(
\m x_i)}{1},
\end{equation}
where $\m y_i \in \mathbb{R}^{N}$ and $\m x_i \in \mathbb{R}^{N}$ are the rasterized groundtruth and input images, while $f\left(\cdot\right)$ is the output of our network. The minimization of the loss function is carried via the Backpropagation Through Time (BPTT) \cite{robinson.2008} algorithm since the weights of the network remain the same for all iterations.

During all our experiments, we used a small batch size of $d=4$ images, the total steps of the network were fixed to K=10 and we set for the initialization of vector $\bm \sigma$ the values $\sigma_{max}=15$ and $\sigma_{min}=1$ . The small batch size is mandatory during training because all intermediate results have to be stored for the BPTT, thus the memory consumption increases linearly to iteration steps and batch size. Furthermore, the optimization is carried again via Adam optimizer and the training starts from a learning rate of $10^{-2}$ which we decrease by a factor of 10 every 30 epochs. Finally, for all trainable parameters we apply $\ell _2$ weight decay of $10^{-8}$.  The full training procedure takes 3 hours for MSR Demosaicking Dataset and 5 days for a small subset of the MIT Demosaicking Dataset on a modern NVIDIA GTX 1080Ti GPU.

\section{Experiments}
\begin{table}[t]
\centering
\begin{tabular}{ccccc}
\hline \hline
\multicolumn{1}{l}{}                                 & Kodak                & McM                  & Vdp                  & Moire                \\ \hline
\multicolumn{1}{l}{\textbf{Non-ML Methods:}}         & \multicolumn{1}{l}{}      & \multicolumn{1}{l}{}    & \multicolumn{1}{l}{}    & \multicolumn{1}{l}{}      \\
bilinear                                             & 32.9                      & 32.5                    & 25.2                    & 27.6                      \\
Adobe Camera Raw 9                                   & 33.9                      & 32.2                    & 27.8                    & 29.8                      \\
Buades \cite{buades2009self}                       & 37.3                      & 35.5                    & 29.7                    & 31.7                      \\
Zhang (NLM) \cite{zhang2011color}                  & 37.9                      & 36.3                    & 30.1                    & 31.9                      \\
Getreuer \cite{getreuer.2011}                      & 38.1                      & 36.1                    & 30.8                    & 32.5                      \\
Heide \cite{heide2014flexisp}                      & 40.0                      & 38.6                    & 27.1                    & 34.9                      \\ \hline
\multicolumn{1}{l}{\textbf{Trained on MSR Dataset:}} &                           &                         &                         &                           \\            
Klatzer \cite{klatzer2016}                         & 35.3                      & 30.8                    & 28.0                    & 30.3                      \\
ours                                                 & 39.2                      & 34.1                    & 29.2                    & 29.7                      \\ \hline
\multicolumn{1}{l}{\textbf{Trained on MIT Dataset:}} & \multicolumn{1}{l}{}      & \multicolumn{1}{l}{}    & \multicolumn{1}{l}{}    & \multicolumn{1}{l}{}      \\
Gharbi \cite{Gharbi:2016:DJD:2980179.2982399}      & 41.2                      & 39.5                    & 34.3                    & \textbf{37.0}           \\
ours*                                                 & \textbf{41.5}           & \textbf{39.7}         & \textbf{34.5}         & \textbf{37.0}           \\ \hline
\end{tabular}
\caption{Comparison of our system to state-of-the-art techniques on the demosaick-only scenario in terms of PSNR performance. The Kodak dataset is resized to 512x768 following the methodology of evaluation in \cite{766768}. $^*$Our system for the MIT dataset was trained on a small subset of 40,000 out of 2.6 million images.}
\label{tbl:general}
\end{table}

Initially, we compare our system to other alternative techniques on the demosaick-only scenario. Our network is trained on the MSR Demosaick dataset \cite{khasabi2014} and it is evaluated on the McMaster \cite{zhang2011color}, Kodak, Moire and VDP dataset \cite{Gharbi:2016:DJD:2980179.2982399}, where all the results are reported in Table \ref{tbl:general}. The MSR Demosaick dataset consists of 200 train images which contain both the linearized 16-bit mosaicked input images and the corresponding linRGB groundtruths that we also augment with horizontal and vertical flipping. For all experiments, in order to quantify the quality of the reconstructions we report the Peak signal-to-noise-ratio (PSNR) metric.

Apart from the MSR dataset, we also train our system on a small subset of 40,000 images from MIT dataset due to the small batch size constraint. Clearly our system is capable of achieving equal and in many cases better performance than the current the state-of-the art network~\cite{Gharbi:2016:DJD:2980179.2982399} which was trained on the full MIT dataset, i.e. 2.6 million images. We believe that training our network on the complete MIT dataset, it will produce even better results for the noise-free scenario. Furthermore, the aforementioned dataset contains only noise-free samples, therefore we don't report any results in Table~\ref{tbl:msr} and we mark the respective results by using N/A instead. We also note that in~\cite{Gharbi:2016:DJD:2980179.2982399}, the authors in order to use the MIT dataset to train their network for the joint demosaicking denoising scenario, pertubed the data by i.i.d Gaussian noise. As a result, their system's performance under the presence of more realistic noise was significantly reduced, which can be clearly seen from Table~\ref{tbl:msr}. The main reason for this is that their noise assumption does not account for the \textit{shot} noise of the camera but only for the \textit{read} noise. 

\begin{table}[!htbp]
\centering
\begin{tabular}{lcccc}
\hline \hline
\multicolumn{1}{l}{}                                        & \multicolumn{2}{c}{noise-free}              & \multicolumn{2}{c}{noisy}                   \\
\multicolumn{1}{l}{}                                        & linRGB               & sRGB                 & linRGB               & sRGB                 \\
\multicolumn{1}{l}{\textbf{Non-ML Methods:}}                &                      &                      &                      &                      \\ \hline
bilinear                                                    & 30.9                 & 24.9                 & -                    & -                    \\
Zhang(NLM) \cite{zhang2011color}                          & 38.4                 & 32.1                 & -                    & -                    \\
Getreuer \cite{getreuer.2011}                             & 39.4                 & 32.9                 & -                    & -                    \\
Heide \cite{heide2014flexisp}                             & 40.0                 & 33.8                 & -                    & -                    \\
\multicolumn{1}{l}{\textbf{Trained on MSR Dataset:}}        & \multicolumn{1}{l}{} & \multicolumn{1}{l}{} & \multicolumn{1}{l}{} & \multicolumn{1}{l}{} \\ \hline
Khasabi \cite{khasabi2014}                                & 39.4                 & 32.6                 & 37.8                 & 31.5                 \\
Klatzer \cite{klatzer2016}                                &  40.9        & 34.6                 & 38.8                 & 32.6                 \\
Bigdeli \cite{bigdeli.2017}                         & -                      & -                    & 38.7                    & -                      \\
ours                                                        & \textbf{41.0}                 & \textbf{34.6}                 & \textbf{39.2}        & \textbf{33.3}        \\
\multicolumn{1}{l}{\textbf{Trained on MIT Dataset:}}        &                      &                      &                      &                      \\ \hline
Gharbi (sRGB)\cite{Gharbi:2016:DJD:2980179.2982399} & 41.6                 & 35.3                 & 38.4                 & 32.5                 \\
Gharbi (linRGB)                                       & \textbf{42.7}                 & \textbf{35.9}                 & 38.6                 & 32.6                 \\
ours* (linRGB)                                         & 42.6                 & \textbf{35.9}        & N/A                  & N/A                  \\ \hline
\end{tabular}
\caption{PSNR performance by different methods in both linear and sRGB spaces. The results of methods that cannot perform denoising are not included for the noisy scenario. Our system for the MIT dataset case was trained on a small subset of 40,000 out of 2.6 million images. The color space in the parentheses indicates the particular color space of the employed training dataset.}
\label{tbl:msr}
\end{table}

Similarly with the noise free case, we train our system on 200 training images from the MSR dataset which are contaminated with simulated sensor noise \cite{foi.2008}. The model was optimized in the linRGB space and the performance was evaluated on both linRGB and sRGB space, as proposed in \cite{khasabi2014}. It is clear that in the noise free scenario, training on million of images corresponds to improved performance, however this doesn't seem to be the case on the noisy scenario as presented in Table \ref{tbl:msr}. Our approach, even though it is based on deep learning techniques, is capable of generalizing better than the state-of-the art system while being trained on a small dataset of 200 images. In detail, the proposed system has a total 380,356 trainable parameters which is considerably smaller than the current state-of-the art~\cite{Gharbi:2016:DJD:2980179.2982399} with 559,776 trainable parameters.

Our demosaicking network is also capable of handling non-Bayer patterns equally well, as shown in Table \ref{tbl:xtrans}. In particular, we considered demosaicking using the Fuji X-Trans CFA pattern, which is a 6x6 grid with the green being the dominant sampled color. We trained from scratch our network on the same train-set of MSR Demosaick Dataset but now we applied the Fuji X-Trans mosaick. In comparison to other systems, we manage to surpass state of the art performance on both linRGB and sRGB space even when comparing with systems trained on million of images. 

On a modern GPU (Nvidia GTX 1080Ti), the whole demosaicking network requires 0.05 sec for a color image of size $220 \times 132$ and it scales linearly to images of different sizes. Since our model solely consists of matrix operations, it could also be easily transfered to application specific integrated circuit (ASIC) in order to achieve a substantial execution time speedup and be integrated to cameras. 

\begin{table}[h]
\centering
\begin{tabular}{lcc}
\hline \hline                                                                    & \multicolumn{2}{c}{noise-free} \\
\multicolumn{1}{l}{}                                 & linear        & sRGB          \\
\multicolumn{1}{l}{\textbf{Trained on MSR Dataset:}} &               &               \\ \hline
Khashabi \cite{khasabi2014}                        & 36.9          & 30.6          \\
Klatzer \cite{klatzer2016}                         & 39.6          & 33.1          \\
ours                                                 & \textbf{39.9} & \textbf{33.7} \\
\multicolumn{1}{l}{\textbf{Trained on MIT Dataset:}} &               &               \\ \hline
Gharbi \cite{Gharbi:2016:DJD:2980179.2982399}      & 39.7          & 33.2   \\ \hline      
\end{tabular}
\caption{Evaluation on noise-free linear data with the non-Bayer mosaick pattern Fuji XTrans.}
\label{tbl:xtrans}
\end{table}
\section{Conclusion}
In this work, we presented a novel deep learning system that produces high-quality images for the joint denoising and demosaicking problem. Our demosaick network yields superior results both quantitative and qualitative compared to the current state-of-the-art network. Meanwhile, our approach is able to generalize well even when trained on small datasets, while the number of parameters is kept low in comparison to other competing solutions. As an interesting future research direction, we plan to explore the applicability of our method on other image restoration problems like image deblurring, inpainting and super-resolution where the degradation operator is unknown or varies from image to image.

\begin{figure}
\begin{tikzpicture}[spy using outlines={circle,yellow,magnification=13,size=2cm, connect spies}]
\node {\pgfimage[width=35mm]{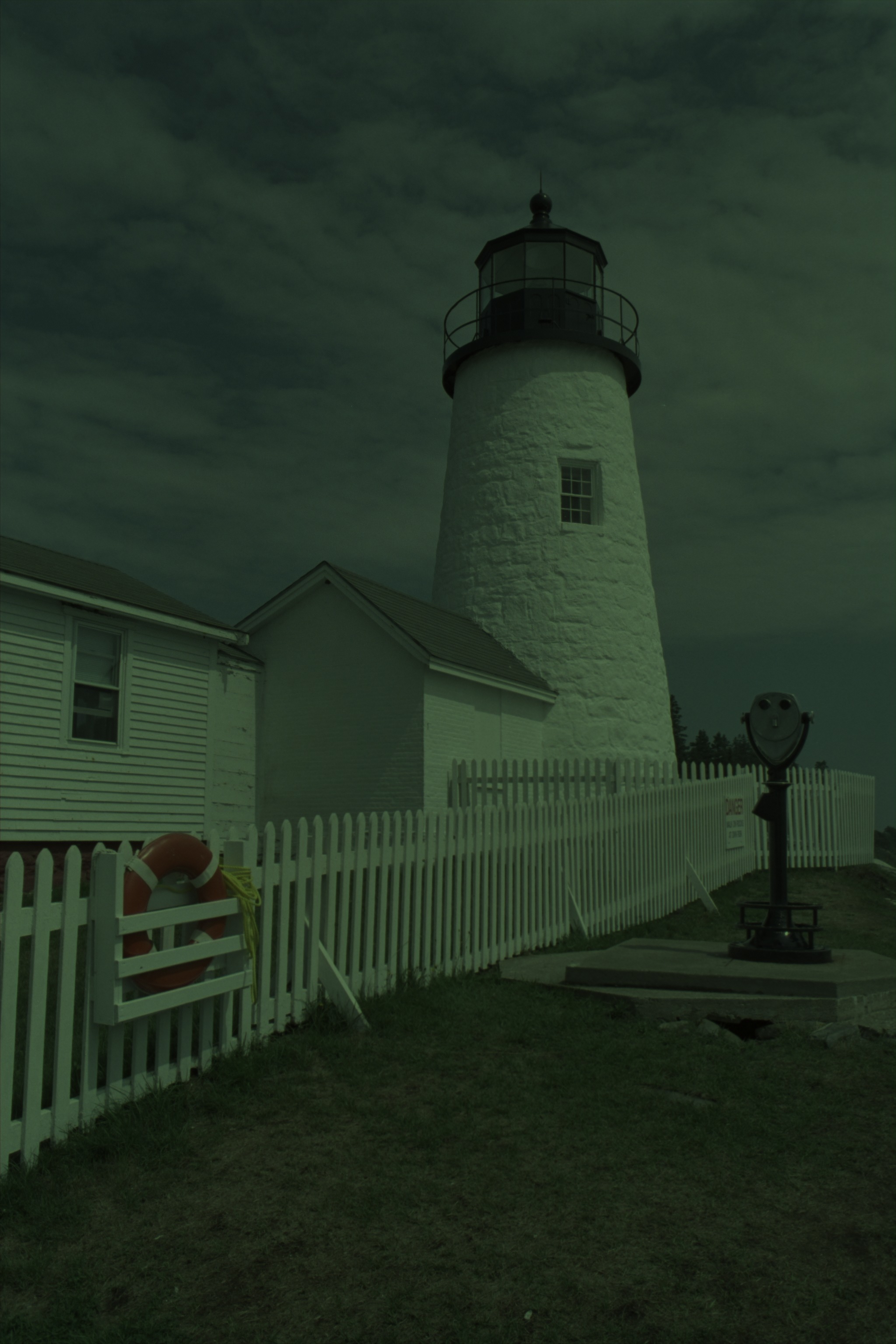}};
\spy on (0.8,-0.8) in node [left] at (0, -2.25);
\end{tikzpicture}
\begin{tikzpicture}[spy using outlines={circle,yellow,magnification=13,size=2cm, connect spies}]
\node {\pgfimage[width=35mm]{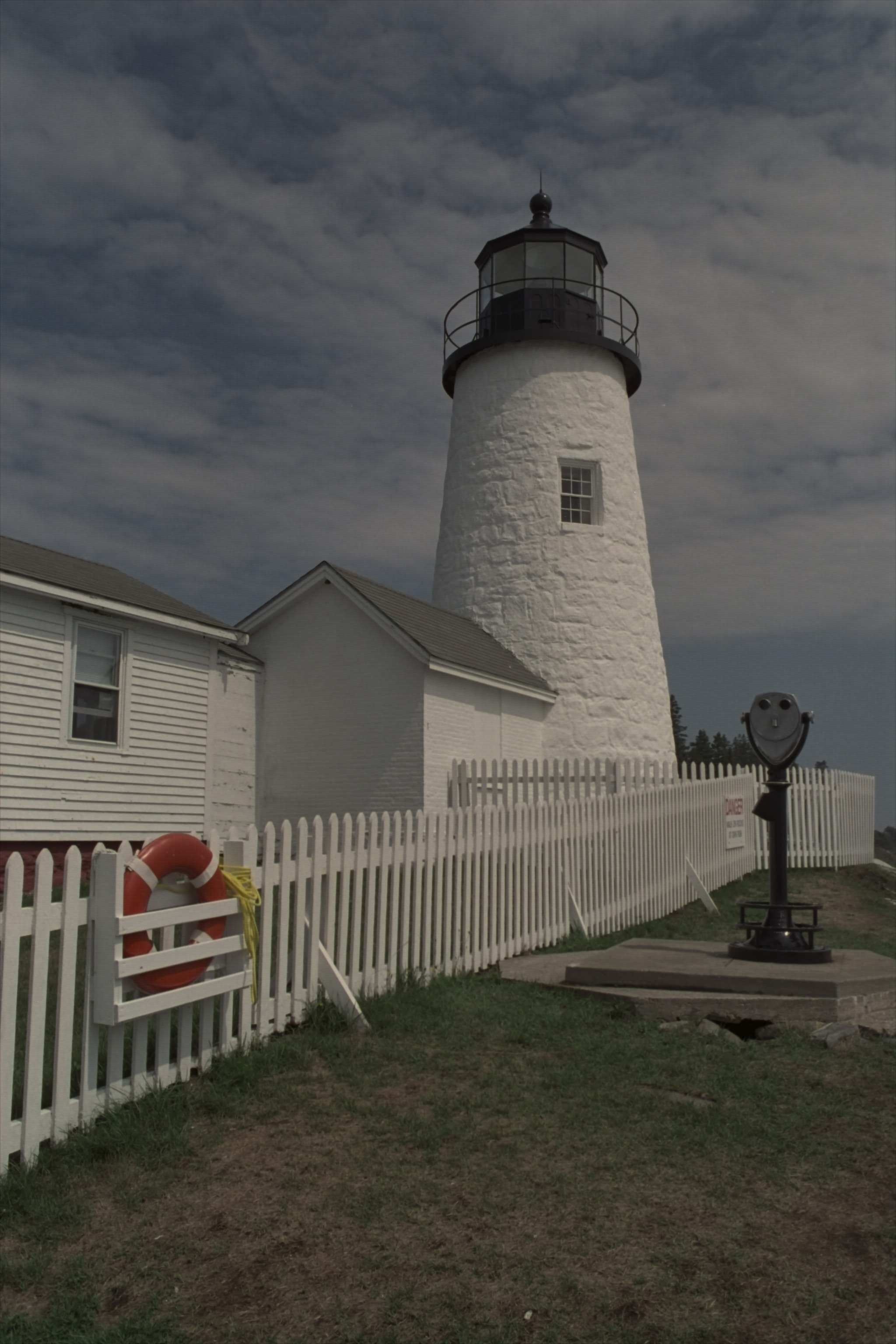}};
\spy on (0.8,-0.8) in node [left] at (0, -2.25);
\end{tikzpicture}
\begin{tikzpicture}[spy using outlines={circle,yellow,magnification=13,size=2cm, connect spies}]
\node {\pgfimage[width=35mm]{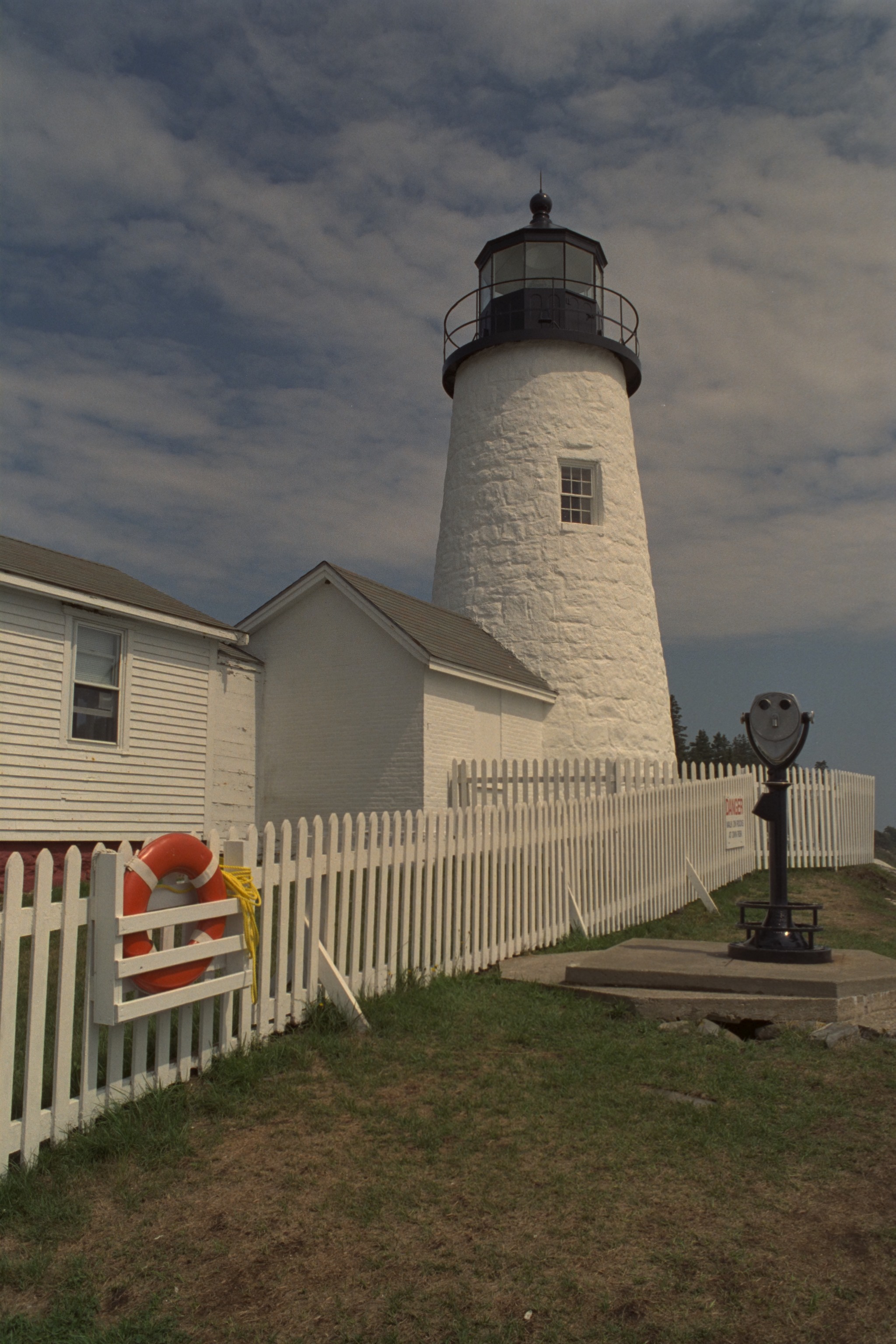}};
\spy on (0.8,-0.8) in node [left] at (0, -2.25);
\end{tikzpicture}

\caption{Progression along the steps of our demosaick network. The first image which corresponds to Step 1 represents a rough approximation of the end result while the second (Step 3) and third image (Step 10) are more refined. This plot depicts the continuation scheme of our approach.} 
\end{figure}

\begin{figure}[!htbp]
  \centering
  \subfloat{\includegraphics[width=0.2\textwidth]{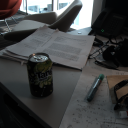}}
  \subfloat{\includegraphics[width=0.2\textwidth]{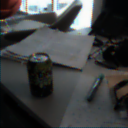}}
  \subfloat{\includegraphics[width=0.2\textwidth]{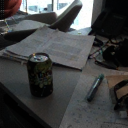}}
  \subfloat{\includegraphics[width=0.2\textwidth]{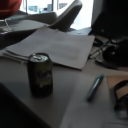}}
  \subfloat{\includegraphics[width=0.2\textwidth]{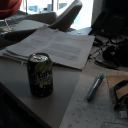}} \\
%   \vspace{-0.03\textwidth} 
%   \subfloat{\includegraphics[width=0.2\textwidth]{images/112/ours_112_original_cropped.png}}
%   \subfloat{\includegraphics[width=0.2\textwidth]{images/112/bi_112_cropped.png}}
%   \subfloat{\includegraphics[width=0.2\textwidth]{images/112/nlm_112_cropped.png}}
%   \subfloat{\includegraphics[width=0.2\textwidth]{images/112/gharbi_noise=0_04_112_cropped.png}}
%   \subfloat{\includegraphics[width=0.2\textwidth]{images/112/ours_112_output_cropped.png}} \\
  \vspace{-0.03\textwidth}
  \subfloat{\includegraphics[width=0.2\textwidth]{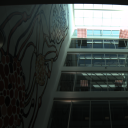}}
  \subfloat{\includegraphics[width=0.2\textwidth]{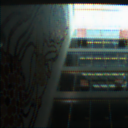}}
  \subfloat{\includegraphics[width=0.2\textwidth]{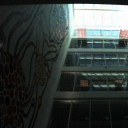}}
  \subfloat{\includegraphics[width=0.2\textwidth]{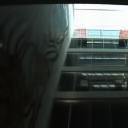}}
  \subfloat{\includegraphics[width=0.2\textwidth]{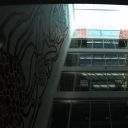}} \\
  \vspace{-0.03\textwidth}
  \subfloat[Reference]{\includegraphics[width=0.2\textwidth]{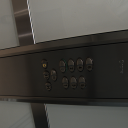}}
  \subfloat[Bilinear]{\includegraphics[width=0.2\textwidth]{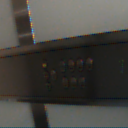}}
  \subfloat[Zhang(NLM)]{\includegraphics[width=0.2\textwidth]{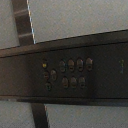}}
  \subfloat[Gharbi et al.]{\includegraphics[width=0.2\textwidth]{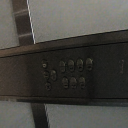}}
  \subfloat[ours]{\includegraphics[width=0.2\textwidth]{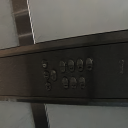}}
  \vspace{-0.03\textwidth}
  \caption{Comparison of our network with other competing techniques on images from the noisy MSR Dataset. From these results is clear that our method is capable of removing the noise while keeping fine details.On the contrary, the rest of the methods either fail to denoise or they oversmooth the images.}
  \label{fig:images}
\end{figure}

\bibliographystyle{splncs}
\bibliography{egbib}
\end{document}